\pdfoutput=1

\documentclass[11pt]{article}

\usepackage{EMNLP2023}

\usepackage{times}
\usepackage{latexsym}

\usepackage[T1]{fontenc}

\usepackage[utf8]{inputenc}

\usepackage{microtype}

\usepackage{inconsolata}
\usepackage{graphicx}

\title{Contextualizing the Limits of Model \& Evaluation Dataset Curation on Semantic Similarity Classification Tasks}

\author{Daniel Theron \\
  Google \\
  \texttt{dtheron@google.com} \\
}

\begin{document}
\maketitle
\begin{abstract}
This paper demonstrates how the limitations of pre-trained models and open evaluation datasets factor into assessing the performance of binary semantic similarity classification tasks. As (1) end-user-facing documentation around the curation of these datasets and pre-trained model training regimes is often not easily accessible and (2) given the lower friction and higher demand to quickly deploy such systems in real-world contexts, our study reinforces prior work showing performance disparities across datasets, embedding techniques and distance metrics, while highlighting the importance of understanding how data is collected, curated and analyzed in semantic similarity classification.
\end{abstract}

\section{Introduction}

With the recent popularization of transformer-based Large Language Models (LLMs) \cite{google_trends_2023} there has been a renewed interest in text embeddings (learned vector representations of words or sentences) for applications including search, recommendations and semantic similarity ranking or classification. At the same time, platforms like TensorFlow Hub, Huggingface and Kaggle have democratized access to a preponderance of pre-trained models tuned for a variety of tasks. In both cases documentation regarding the training, tuning and limits of models is often incomplete or difficult to find / consume, with an increased incentive and velocity towards quick deployment of these systems in real-world contexts.

In this paper we investigate the performance of a binary semantic similarity classification task to demonstrate these challenges. Though a strict formalization of the term \textit{semantic similarity} is an open problem across a number of disciplines, we use the term in its colloquial machine learning sense: as the degree of likeness in meaning between texts, rather than the more general concept of semantic relatedness, which includes lexical relationships that may drastically alter the likeness in meaning \cite{Budanitsky2004SemanticDI}.

 We perform this study using a variety of accessible pre-trained models on 3 commonly used evaluation datasets, reflecting on the provenance, characteristics and limitations of both models and datasets as documented and measured by the original authors or related works. We consider the impact of these factors, as well as the choice of distance metric on the classification task. In doing so we hope to reinforce that there exist notable differences in model performance across these datasets within the context of the ethical and architectural considerations of the artifacts themselves.

\begin{table*}
\centering
\begin{tabular}{lll}
\hline
\textbf{Entity} & \textbf{Models} & \textbf{Datasets}\\
\hline
    \bf Cards (M) & & \\
      \hspace{3mm} Download-Weighted & 0.852 ($\pm 0.044$) & 0.848 ($\pm 0.0445$)  \\
      \hspace{3mm} Uniform           & 0.552 ($\pm 0.0616$)& 0.58 ($\pm 0.0612$)\\
    \bf Cards with Disclosures (K) & & \\
      \hspace{3mm} Download-Weighted & 0.508 ($\pm 0.062$) & 0.716 ($\pm 0.0559$)  \\
      \hspace{3mm} Uniform           & 0.192 ($\pm 0.0488$)& 0.16 ($\pm 0.0454$)\\
\hline
\end{tabular}
\caption {Huggingface Data and Model Card Prevalence} \label{table:3} 
\end{table*}

\section{Ethical Considerations}

As we discuss in the Datasets, Models and Embeddings sections of this paper, both the evaluation datasets we use, as well as the pretrained language models' training data skew heavily English (in some cases by design) and Western (in terms of the web platforms from which data was originally sourced, along with the user demographics of said platforms). Further, given historical gender divides in access to internet, mobile and digital technologies, as well as cultural barriers faced by women (particularly in education and labor participation) \cite{oecd_gender_2018} we presume a meaningful male bias in these corpus' web content.

As we use these datasets to evaluate semantic similarity classification on associated English text embeddings produced by these pretrained models, we therefore expect our study results to be largely relevant only to contexts that conform to these demographics. In part, our work in this paper to understand the origin, curation methods and contours of datasets and models serves to highlight the need to constrain conclusions about machine learning task performance based on the limits of their data as much as their architectures, objectives and parameters. 

Forward-looking work by Mitchell et al. on model cards and by Pushkarna et al. on data cards have suggested ways to standardize these types of disclosures \cite{Mitchell_2019, pushkarna2022data}. As a benchmark on the accessibility of such disclosures we selected Huggingface as an exemplar for both ease-of-use and their focus on making model cards both easier for developers to produce, and easier for end-users to consume \cite{hf_model_Cards_2022}. We drew four samples ($n=250$) from their platform on May 16, 2023: both for model and data cards, both download-weighted and uniformly, and calculated a binomial confidence interval at $\alpha = 0.05$ to estimate $M$ the prevalence of model cards (conservatively, as the existence of a README file as per the Huggingface model card guide) and $K$ the prevalence of model cards \emph{with} terms associated with fairness, citations, annotators or limitations (see Table \ref{table:3}). Though coarse, these metrics may suggest significant further work is needed to widen the reach and depth of disclosure.

In all, these observations motivate caution in the deployment of such systems without increased due diligence - especially in consequential domains - given the potential for disparate impact on users from out-of-distribution groups.

\section{Methods}

\subsection{Datasets}

We used the Quora Question Pairs (QQP), Microsoft Research Paraphrase Corpus (MRPC) and the Semantic Textual Similarity Benchmark (STSB) datasets by way of the General Language Understanding Evaluation benchmark (GLUE) collection hosted on the Huggingface platform to perform benchmarking \cite{wang2019glue}. The datasets are all in English and contain pairs of sentences and associated ground truth labels indicating whether the sentences are similar or dissimilar from one another. The mean number of words per sentence is 12 ($\sigma=6$). While QQP and MRPC labels are binary, STSB labels follow a Likert scale from 0-5 (with 5 being exactly the same).

\begin{table*}
\centering
\begin{tabular}{lll}
\hline
\textbf{Dataset / Similar} & \textbf{Sentence Pairs}\\
\hline
  \bf QQP & \\
    \hspace{3mm} False & what does it mean if you keep dreaming about someone else being pregnant  \\
    \hspace{3mm} & what does it mean if i dream im pregnant \\
    \hspace{3mm} True & are we all hypocrites really \\
    \hspace{3mm} & are we all hypocrites justify \\
  \bf MRPC & \\
    \hspace{3mm} False & the european union banned the import of genetically modified food in 1998 \\
    \hspace{3mm} & the united states is now demanding that the eu end its ban \\
    \hspace{3mm} & the union banned the import of genetically modified food in 1998 after many \\
    \hspace{3mm} & consumers feared health risks \\
    \hspace{3mm} True & general jeffrey said he would donate his military pension to charity for the period \\
    \hspace{3mm} & he was in office at yarralumla \\
    \hspace{3mm} & majgen jeffery said he would give his military pension to charity while he served \\
    \hspace{3mm} & at yarralumla \\
  \bf STSB & \\
    \hspace{3mm} False & china stocks close mixed friday \\
    \hspace{3mm} & chinese stocks close higher midday friday \\
    \hspace{3mm} True & a puppy is sliding backwards along the floor \\
    \hspace{3mm} & a puppy is pushing itself backwards \\
\hline
\end{tabular}
\caption {Sample Sentence Pairs} \label{table:1} 
\end{table*}

To better understand the relationship between each dataset's ground truth labels and structural features of associated sentence pairs we computed several sentence metrics ("study features") that formed the basis of downstream analysis, including:

\begin{itemize}
  \item \textbf{Pairwise Levenshtein Distance.} Calculates the minimum number of single-character transformations required to change one string into another. The metric may be interpreted as the degree of character-level difference between two strings. \cite{levenshtein_1966}.
  \item \textbf{Mean Dale-Chall Score.} A metric for calculating reading comprehension difficulty based on a list of 3,000 words that American 4th grade students might reasonably understand. The list was originally published in 1948 and updated in 1995. The score itself uses a 10-point scale, with bins corresponding to various grade-level proficiencies and is used to represent the comprehension difficulty of a text \cite{chall_1948} \cite{chall1995readability}. As many sentence pairs in this study are below the lower threshold of 100 words used by the updated Dale-Chall formula, we padded shorter strings with an in-corpus word prior to calculating the metric. As such our measure of Dale-Chall may more accurately be described as a weighted Dale-Chall score.
  \item \textbf{Mean Type Token Ratio (TTR).} A measure of vocabulary variation, TTR is the ratio between the number of unique types (words) in a text and the total number of words. It is interpreted as describing the lexical density of a text \cite{ure_1971}.
  \item \textbf{Sentence Vocabulary Intersection}. The number of vocabulary words common between sentences in a pair.
  \item \textbf{Word Synset Intersection.} A metric we use to interpret the overlap in the comprehension space between sentences. Using each word in a sentence's vocabulary, we extract synonyms from WordNet's first order synset / sense \cite{wordnet_2010}. We then count the number of synonyms common between sentences in a pair.
  \item \textbf{Parts of Speech, Character \& Word Counts.} Counts including the mean number of words and characters in a sentence pair, as well as the number of verbs, nouns, etc. in a pair. Part of speech classification was done with NLTK using the universal tagset \cite{ntlk_2023}.
\end{itemize}

We fit LightGBM classifiers to these features ($AUC\approx0.8$), predicting binary label values, and used Shapley Additive Explanations (SHAP) \cite{shap_2017} to understand the contribution of various features to sentence similarity. We also measured the Point Biserial Correlation (PBC) at $\alpha=0.05$ between the labels and features (see Table \ref{table:2}).

Our analysis suggests that both QQP and STSB datasets predominantly feature sentence pairs with lower structural complexity, and that even within the MRPC dataset there is a stronger relationship between simpler sentences and ground truth labels.  These observations, in conjunction with prior work (discussed in the subsections below) suggesting unknown label provenance, known label instability and ambiguity in the proficiencies and demographic distribution of raters may be useful for data science practitioners to keep in mind when using them to train or evaluate semantic similarity classifiers for broader, out-of-data-context applications.

\begin{table*}
\centering
\begin{tabular}{lrrr}
\hline
         \bf Sentence Feature &    \bf QQP &   \bf MRPC &   \bf STSB \\
\hline
   Num. Overlapping Words & 0.2065 & 0.3338 & 0.4192 \\
Num. Overlapping Synonyms & 0.2018 & 0.2812 & 0.3793 \\
         Type-Token Ratio & 0.1241 &     &     \\
                     VERB &     & 0.1921 &     \\
Mean Sentence Num. Chars. &     &     & 0.2446 \\
 Mean Sentence Num. Words &     &     & 0.2349 \\
                     NOUN &     &     & 0.2142 \\
\hline
\end{tabular}
\caption {PBC for Top N Sentence Features} \label{table:2} 
\end{table*}

\subsubsection{QQP}

The QQP dataset was released in 2017 by Quora, the crowdsourced question answering website. The dataset consists of 400,000 question pairs \cite{quora_2016}. While we were not able to find any detailed discussion from Quora around how pairs were labeled, their 2017 Kaggle competition description makes mention of using a random forest model in production \cite{kaggle_2017}. This might presuppose the existence and continued maintenance of training data with verified ground truth labels for supervised or weakly supervised learning. In fact, at least as of 2014 we know Quora does operate a content moderation team that might also be involved in broader data labeling activities \cite{quora_2014}.

We therefore assume that QQP labels may represent some unknown combination of automated and human decisions. Dadashov et al. performed a blind study on 200 sampled rows from QQP as part of their paper comparing various semantic similarity methods, in which they measured the agreement rate in classification decisions between the original Quora label and an independent reviewer. They found an 83.5\% match rate between the raters, indicating some degree of label instability \cite{dadashov_2017}. This conforms with Quora's own disclaimer around label quality \cite{quora_2016}.

\begin{figure}[h]
  \caption{QQP Study Features' Shapley Values} \label{figure:1} 
  \centering
  \includegraphics[scale=0.5]{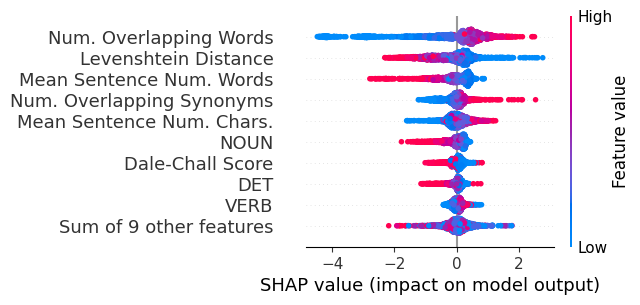}
\end{figure}

Our analysis of the study features' PBC coefficients suggest that higher overlap in vocabulary between sentence pairs, as well as higher overlap in synset intersections are correlated with the model's predictions (binary sentence similarity) with $PBC>0.201$ (see Table \ref{table:2}). SHAP values for a LightGBM classifier reinforce PBC conclusions (see Figure \ref{figure:1}). Further, SHAP values suggest that fewer words and lower Levenshtein distances between sentence pairs also contribute to model predictions. Considering a sample of QQP sentence pairs (see Table \ref{table:1}), we note that questions are typically short and direct, with 73\% of pairs at a high school reading level or below (as per Dale-Chall).

\subsubsection{MRPC}

MRPC was published in 2005 and contains 5,801 sentence pairs selected via heuristics and Support Vector Machine (SVM) from a topic-clustered pool of news data. Ground truth labels are binary and represent whether two raters (or in case of ties, three) considered each pair semantically equivalent. As the authors note, they had to relax their rubric for "semantically equivalent" from strict symmetrical entailment in order to produce a corpus more robust than virtually identical string pairs. Perhaps an artifact of the more loose rating criteria, the authors noted an 84\% inter-rater agreement at $\kappa=62$ \cite{Dolan2005AutomaticallyCA}.

\begin{figure}[h]
  \caption{MRPC Study Features' Shapley Values} \label{figure:2} 
  \centering
  \includegraphics[scale=0.5]{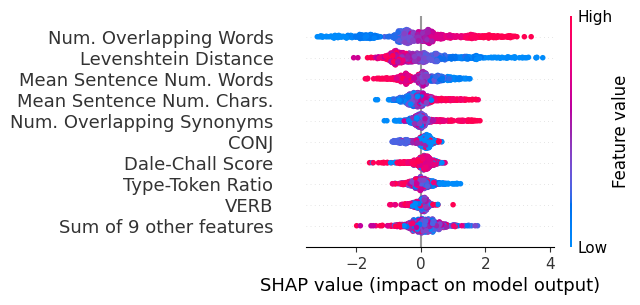}
\end{figure}

Here, measured $PBC>0.2812$ also indicates that both a greater overlap in vocabulary, and synset intersections between sentence pairs is correlated with ground truth labels (see Table \ref{table:2}). Similar to QQP, SHAP values show lower Levenshtein distances have a higher impact on model predictions, however for MRPC fewer words and more characters in sentence pairs also contribute to label values (see Figure \ref{figure:2}). This could suggest that raters were likely to rate as similar shorter sentence pairs, pairs with longer words, as well as pairs with words repeated (either directly or as synonyms of one another) across both sentences. Investigating sample sentence pairs from MRPC (see Table \ref{table:1}) we note that the dataset often contains names, jargon and numbers; factors that may explain the modeled raters' propensity to anchor similarity classification on these particular structural features. MRPC sentence pairs also have the highest Dale-Chall scores of the study, with more than 88\% scoring at a college reading level or above.

\subsubsection{STSB}

STSB was published in 2017 as part of the SemEval-2017 Task 1 workshop on semantic textual similarity methods and is made up of selected English sentence pairs from SemEval tasks from 2012 - 2017 \cite{cer-etal-2017-semeval}. Data sources include news articles, image captions and forum posts. Labels were crowdsourced using Amazon's Mechanical Turk service, with five annotations collected per pair and averaged to produce ground truth ratings on a scale of 0 - 5. Cer et al. neither mention details around any inter-rater agreement or quality assurance procedures they performed during the labeling process, nor provide data regarding the distribution of rater demographics.

We chose to binarize STSB labels around $score >= 3$. This was motivated by the SemEval annotation rubric, which defines this threshold as "... sentences... roughly equivalent, but some important information differs..." \cite{cer-etal-2017-semeval}. Further, we noted that the mean sentence vocabulary intersection is consistently above 6 words at $score >= 3$ (less than 4 words below), with $PBC=0.4192$ with respect to ground truth labels.

\begin{figure}[h]
  \caption{STSB Study Features' Shapley Values} \label{figure:3} 
  \centering
  \includegraphics[scale=0.5]{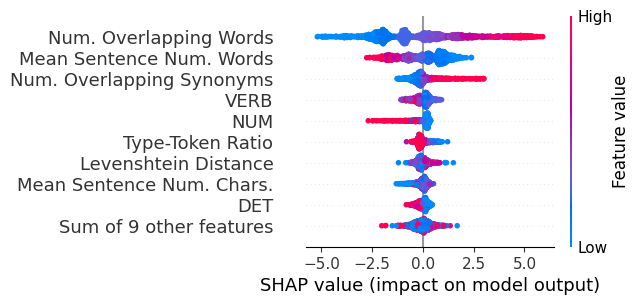}
\end{figure}

Inspection of the study features revealed that in addition to sentence vocabulary intersection, word synset intersection was also reasonably correlated with labels at $PBC=0.3793$ (see Table \ref{table:2}). SHAP values indicate the same, and also show that a smaller sentence vocabulary contributes to model predictions (see Figure \ref{figure:3}). Unlike QQP and MRPC, STSB SHAP values also suggest that a lower number of verbs in sentence pairs is within the top 4 study features with respect to predictive power. These observations appear to align with Dale-Chall scores, which place 65\% of sentence pairs at or below a high-school reading level.

\subsection{Models}

We encoded each dataset's sentence pairs using  Huggingface's ALL-MPNET-BASE-V2, Sanh et al.'s DISTILROBERTA-BASE, Google's Language-agnostic BERT sentence embedding model (LaBSE), and OpenAI's TEXT-EMBEDDING-ADA-002 model. We subsequently used these text embedding pairs in combination with a variety of distance metrics to evaluate sentence similarity classification performance. We also directly classified a sample of sentence pairs from each dataset using OpenAI's GPT-4 API.

These models' training data were all largely sourced from the internet (including, we speculate based on press releases from OpenAI, the proprietary TEXT-EMBEDDING-ADA-002 and GPT-4). Given that the QQP, MRPC and STSB datasets predate most models, there is a significant risk that these datasets themselves may have been included in the study models' training data. As our study attempts to simulate how models' ability to generalize across unseen examples are evaluated in practice, the risk of data contamination should give practitioners pause as to the validity of their evaluation metrics. In fact, ALL-MPNET-BASE-V2 explicitly includes QQP in its training corpus \cite{song2020mpnet}.

\subsubsection{ALL-MPNET-BASE-V2}

Microsoft introduced MPNet, a combination masked and permuted language modeling approach for language understanding, in 2020. The original model was trained on a 160GB corpus of text, including Wikipedia, web text, news articles and books, and fine tuned on GLUE tasks, the Stanford Question Answering Dataset (SQuAD) task, the ReAding Comprehension from Examinations (RACE) multiple-choice answer task and IMDB sentiment classification task \cite{song2020mpnet}.

Our study used Huggingface's ALL-MPNET-BASE-V2: a fine-tuned version of MPNet using a set of 1B sentence pairs towards a contrastive learning objective over 768-dimensional vectors with a token length of 384 \cite{nreimers_mpnet_2021}. Huggingface's training data consists primarily of Reddit comments, but also includes citation pairs, question and answer pairs, as well as comments with code pairs and image captions. The corpus skews heavily English and Western (both in terms of the web platforms and associated users that originally generated the data) for both the base and Huggingface fine-tuned versions of the model. It has consistently been the most downloaded sentence similarity model on Huggingface, with more than 10M downloads in October, 2023 alone.

\subsubsection{DISTILROBERTA-BASE}

The DistilRoBERTa base model was produced in 2019 by Sanh et al. through distillation of the RoBERTa base transformer, itself having been trained for masked language modeling (MLM) \cite{Sanh2019DistilBERTAD}. RoBERTa base was trained on an all-English corpus of unpublished books, Wikipedia articles, news articles and open web text \cite{DBLP:journals/corr/abs-1907-11692}. The DistilRoBERTa student was trained on open web text and produces 768-dimension text embeddings. Both datasets follow similar content and user demographic patterns as those found in ALL-MPNET-BASE-V2. Similar to that model, DistilRoBERTa base also continues to be among the most downloaded on Huggingface.

\subsubsection{LaBSE}

This model from Feng et al. is based on the BERT transformer and is optimized to generate similar representations for translated bilingual sentence pairs \cite{feng2022languageagnostic}. Of the dual-encoder architecture, an MLM was pre-trained with monolingual CommonCrawl and Wikipedia data, while a translation language model (TLM) was trained with bilingual sentence pairs sourced from web pages with filtering heuristics and limited human annotation. The MLM corpus includes a significant percentage of English, Russian, Japanese, simplified Chinese and French sentences. However, the corpus also contains a long tail of examples from 105 additional languages. LaBSE's bilingual corpus, by contrast, contains roughly equivalent sets of en-xx pairs from 64 of these same languages (also including Hindi, Korean, Swahili, etc.), with more limited examples from the remaining 41 \cite{feng2022languageagnostic}. While not as popular in downloads as the prior study models, LaBSE represents an attempt at producing a cross-lingual semantic similarity embedding model.

\subsubsection{TEXT-EMBEDDING-ADA-002 and GPT-4}

TEXT-EMBEDDING-ADA-002 is a 2nd generation embeddings-as-a-service API endpoint model from OpenAI that was released in 2022 with 1,536 dimensions \cite{openai_ada}. As a proprietary commercial product, little is publicly known about the architecture, training data and model objective of ADA-002 compared to other models in this study. However, with the recent popularization of LLMs and embeddings-as-a-service, we included both ADA-002, as well as OpenAI's conversational LLM GPT-4 as benchmarks for semantic similarity classification in this study.

Importantly, GPT-4 is a chat completion LLM fine tuned with Reinforcement Learning from Human Feedback (RLHF) \cite{openai_gpt4}. Rather than computing distance metrics for semantic similarity classification using model embeddings (as with the other models in this study), we formulate the task for GPT-4 as an English prompt to GPT-4, appending 5 randomly sampled examples from a given training set to the target pair (in-context learning).

\begin{figure*}[h]
  \caption{Semantic Similarity Classification Performance} \label{figure:4} 
  \centering
  \includegraphics[scale=0.5]{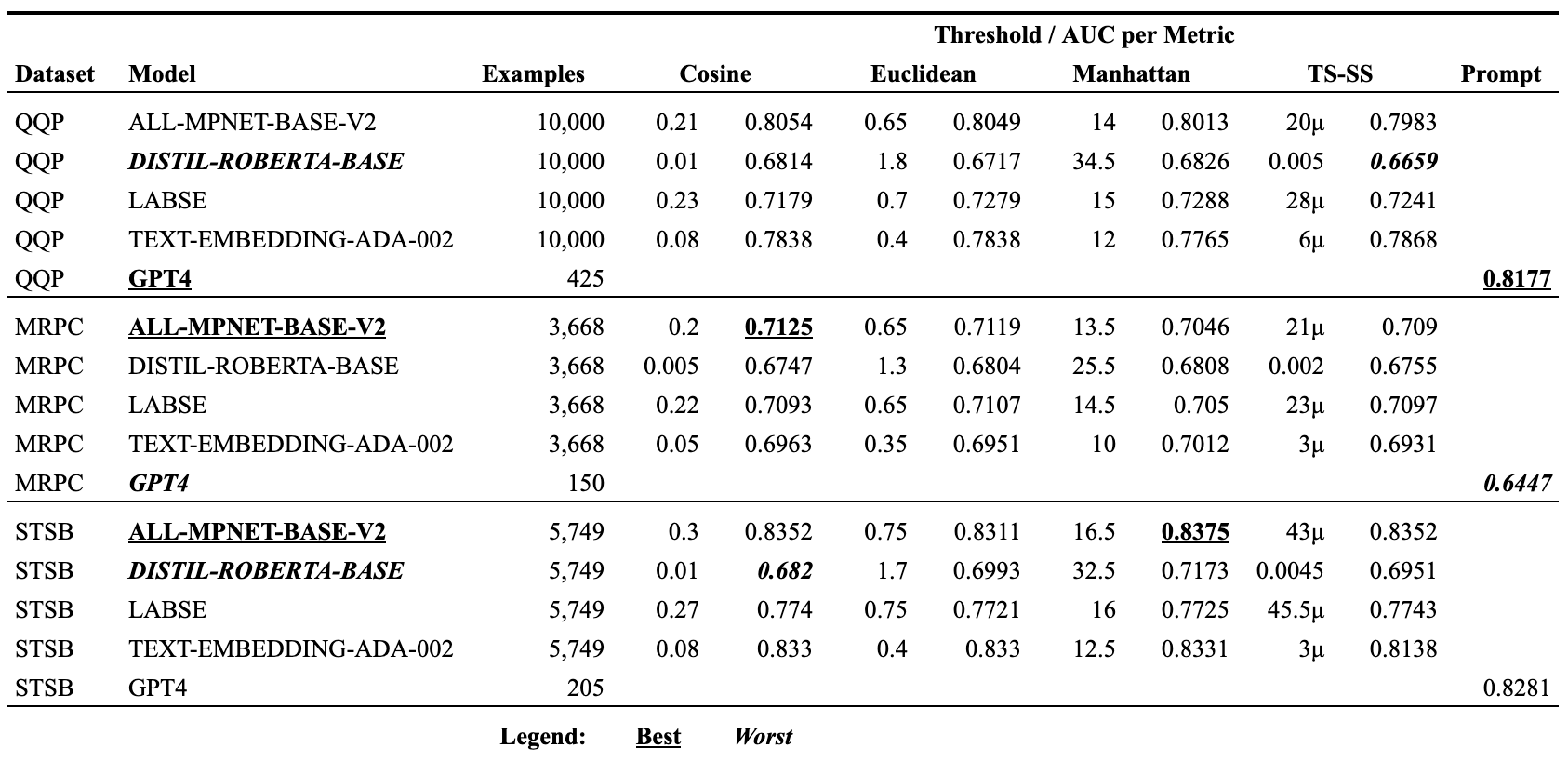}
\end{figure*}

\subsection{Embeddings and Distance Metrics}

We encoded sentences in each study dataset with each study model (except GPT-4; see below) to produce text embedding pairs. Then, within each dataset, we calculated the distance between each pair of vectors using cosine, euclidean, manhattan and Triangle Area Similarity / Sector Area Similarity (TS-SS) metrics. We performed a gridsearch across metric classification thresholds, calculating the True Positive Rate (TPR), False Positive Rate (FPR) and Area Under the Curve (AUC) for each with respect to ground truth labels.

For GPT-4, we sent a sample of plain text sentence pairs to the OpenAI GPT-4 API via the following prompt, including 5 random examples from the associated dataset's ground truth for in-context learning:

\begin{quote}
\emph{Are the following two sentences semantically similar to each other? Respond only with a 1 if they are similar and a 0 if they are not. Here are a few examples:}

\emph{Random pair sentence 1.\\
Random pair sentence 2.\\
Output: 1}\\

\emph{Random pair sentence 1.\\
Random pair sentence 2.\\
Output: 0}\\

...\\

\emph{Target pair sentence 1.\\ 
Target pair sentence 2.\\
Output:}
\end{quote}

We calculated the same TPR, FPR and AUC metrics for these samples.

\section{Results \& Conclusions}

Overall, ALL-MPNET-BASE-V2 proved robust across all 3 study datasets, with the best AUC on MRPC and STSB similarity classification (0.7125 and 0.8375 respectively). While GPT-4 was the best performing model on QQP dataset (0.8177), it's AUC was also the most unstable across datasets, with a 0.1834 point spread between STSB and MRPC. Though the choice of distance metric produced marginal differences in score, the effect of model choice and dataset were more pronounced. 

Models generally performed better on STSB with mean $AUC = 0.7837$ across metrics and models, and the worst on MRPC at mean $AUC = 0.6981$ (see Figure \ref{figure:4}).

Together these results align with the observations regarding dataset and ground truth curation in the Datasets section. Specifically, the higher complexity of sentences in MRPC, along with measured inter-rater agreement of only 85\% may go towards explaining why our study showed the worst performance across models on this dataset. At the same time, both STSB and QQP sentences displayed more structural simplicity, and while there is some documented inter-rater disagreement in at least one of these, this may explain generally better study performance across both.

Further, given the limits of the study (English-only), as well as the model and dataset content, ground truth labels, and labeling processes discussed in the Ethical Considerations, Datasets and Models sections, caution should be taken in the deployment of these pretrained models / evaluation criteria for semantic similarity classification in real-world contexts. 

\bibliography{emnlp2023}
\bibliographystyle{acl_natbib}

\end{document}